\documentclass[letterpaper, 10 pt, conference]{ieeeconf}  % Comment this line out if you need a4paper

\IEEEoverridecommandlockouts                              % This command is only needed if 
\usepackage{graphicx} % for pdf, bitmapped graphics files
\usepackage{amsmath} % assumes amsmath package installed
\usepackage{bm}    
\usepackage{amssymb}  % assumes amsmath package installed
\usepackage{caption}
\usepackage{subcaption}
\usepackage{float}
\usepackage[ruled,vlined]{algorithm2e}

\usepackage{color}

\newcommand{\RV}[1]{{\color{black}#1}}
\title{\LARGE \bf
Learning Variable Impedance Control via Inverse Reinforcement Learning for Force-Related Tasks
}

\author{Xiang Zhang$^{1}$, Liting Sun$^{1}$, Zhian Kuang$^{1,2}$ and Masayoshi Tomizuka$^{1}$% <-this % stops a space
\thanks{$^{1}$ Xiang Zhang, Liting Sun, and Masayoshi Tomizuka are with the Department of Mechanical Engineering, University of California, Berkeley, CA 94720, USA.
{\tt\small xiang\_zhang\_98@berkeley.edu, litingsun@berkeley.edu, tomizuka@berkeley.edu}}%
\thanks{$^{2}$ Zhian Kuang is with the Research Institute of Intelligent Control and Systems, Harbin Institute of Technology, 150001, Harbin, P.R. China. {\tt\small zhiankuang@berkeley.edu}}%

\thanks{$^{1}$
Zhian Kuang is a visiting scholar with the Department of Mechanical Engineering, University of California, Berkeley, CA 94720, USA.}
}

\begin{document}

\maketitle
\thispagestyle{empty}
\pagestyle{empty}

%%%%%%%%%%%%%%%%%%%%%%%%%%%%%%%%%%%%%%%%%%%%%%%%%%%%%%%%%%%%%%%%%%%%%%%%%%%%%%%%
\begin{abstract}
% Impedance control, as a position-force control method, has been widely applied on robot systems to ensure safe robot interactions with unknown environment. Recently, more and more complex manipulation tasks require robots to adapt impedance according to the task phases and designing and tuning such variable impedance laws is nontrivial. To obtain the variable control skill, many learning based method have been utilized to 

Many manipulation tasks require robots to interact with unknown environments. In such applications, the ability to adapt the impedance according to different task phases and environment constraints is crucial for safety and performance. Although many approaches based on deep reinforcement learning (RL) and learning from demonstration (LfD) have been proposed to obtain variable impedance skills on contact-rich manipulation tasks, these skills are typically task-specific and could be sensitive to changes in task settings. This paper proposes an inverse reinforcement learning (IRL) based approach to recover both the variable impedance policy and reward function from expert demonstrations. We explore different action space of the reward functions to achieve a more general representation of expert variable impedance skills. Experiments on two variable impedance tasks (Peg-in-Hole and Cup-on-Plate) were conducted in both simulations and on a real FANUC LR Mate 200iD/7L industrial robot. The comparison results with behavior cloning and force-based IRL proved that the learned reward function in the gain action space has better transferability than in the force space. Experiment videos are available at https://msc.berkeley.edu/research/impedance-irl.html.
\end{abstract}

% \copyrightnotice

%%%%%%%%%%%%%%%%%%%%%%%%%%%%%%%%%%%%%%%%%%%%%%%%%%%%%%%%%%%%%%%%%%%%%%%%%%%%%%%%
\section{INTRODUCTION}

% Robot systems are increasingly being deployed into unstructured environments, where they are expected to interact with the unmodeled environments and perform complex tasks. To achieve safe interactions with the environment, impedance control has been applied on such robot systems. The basic idea of impedance control is to set  a  virtual  mass-spring-damping  relationship  between  robots  and  the  unknown  environment.
Robot systems are increasingly deployed into various unstructured environments (e.g., factories, houses, hospitals). In such environments, robots are expected to perform complex manipulation tasks while interacting with unknown environments in a safe and stable manner. Impedance control, which establishes a virtual mass-spring-damping contact dynamic, has been widely applied to these robot systems to guarantee safe physical interactions. Moreover, many complex manipulation tasks require the robot to change impedance according to the task phases. In practice, a variable impedance skill is needed in such tasks.

In recent years, several learning-based methods have been introduced to obtain variable impedance skills. Examples include learning from demonstrations (LfD)~\cite{peternel2015human,abu2018force,tang2016teach}, deep reinforcement learning (RL) with variable impedance action spaces~\cite{buchli2011learning,rey2018learning,martin2019variable}. However, the task-specific impedance skills obtained by LfD approaches may fail when the task changes. Besides, designing a suitable reward function is challenging for RL. Therefore, their skill transferability is limited. A more general way of learning variable impedance skills needs to be found to tackle these problems.

\begin{figure}
    \centering
    \includegraphics[scale=0.4]{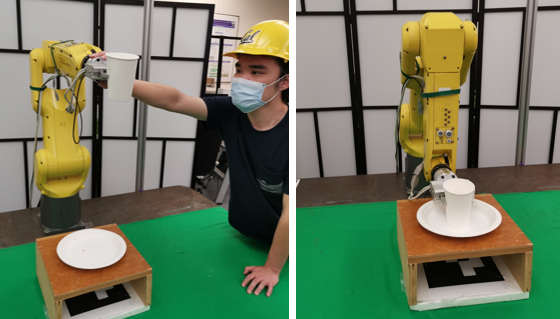}
    \caption{Left: a human expert demonstrates on the robot, Right: the robot accomplishes the task with the learned variable impedance skill}
    \label{fig:teaching}
\end{figure}
% \LS{The goal of this work is to explore a better format of learning the force-related tasks in robots. Hence, you should mention your contribution in terms of comparing the results with gain policy and the force policy.}

The development of the inverse reinforcement learning (IRL) method provides new insight on learning from demonstration. The goal of IRL is to recover the expert reward or cost function from demonstrations. New policies can then be obtained by reoptimizing this learned reward in new scenarios. Previously, IRL has been applied to several tasks such as plate placing~\cite{finn2016guided} and route planning~\cite{ziebart2008maximum}. However, there is no previous work applying the IRL method to the variable impedance control to our knowledge. Furthermore, the action space for the reward function for this task is still unclear. Should we define rewards directly on the force, or should we look for a reward function in terms of the impedance gain?

This paper proposes an inverse reinforcement learning based approach to recover both the variable impedance policy and the reward function from expert demonstrations. While this learned policy can be utilized to solve the original task, new variable impedance policies can be generated for different task settings \RV{by using RL to maximize the learned reward function. Since the learned reward function only depends on the current observation and action, it is agnostic to task settings.} Thus, our approach is more general and has better transferability in comparison with previous LfD approaches.  

Furthermore, similar to~\cite{martin2019variable}, we argue that, for force-related tasks, learning in the impedance gain action space is better than learning in the force action space. The basic idea is that the performance of gain policy is guaranteed by the impedance control law and therefore improves the reward transferability. For validation, we compared our approach with behavior cloning (BC) \cite{pomerleau1989alvinn} baseline in two tasks: Peg-in-Hole and Cup-on-Plate (as shown in Fig.~\ref{fig:teaching}), and the influence of two action spaces (impedance gain and Cartesian space force) have also been studied. As a result, our approach successfully recovers the expert variable impedance policy and achieves better transfer performance than BC and the force-based IRL in the testing scenarios.

This paper's remainder is organized as follows: Related works are introduced in Section \ref{RELATED WORK}. Our controller design and learning framework are outlined in Section \ref{PROPOSED APPROACH}. Section \ref{EXPERIMENTS} presents the experiments in simulation and on the real robot. Finally, the conclusion is given in Section \ref{conclusion}

\section{RELATED WORK}\label{RELATED WORK}
% \LS{you should also talk about robot force-control related tasks and the approaches used.}
\subsection{Force-control related tasks in robotics}
Force control is essential for many robotics applications, such as the assembly, the surface polishing, and the machining. Previous works in this field can be categorized into two methods: pure force control and position-force control. The pure force control accomplishes tasks by directly controlling the robot's force in the Cartesian space or in the joint space. Examples can be found in the peg-hole assembly task~\cite{inoue2017deep} and visual manipulations tasks from~\cite{levine2016end}. For the position-force control, it can be further separated into the hybrid position/force control~\cite{raibert1981hybrid} and the impedance control~\cite{hogan1985impedance}. In the first method, the position and force are controlled in two separate channels. Researchers have deployed this method on surface polishing tasks and peg-in-hole~\cite{tang2016teach}. However, the position-force decoupling can only be achieved when the task is well-defined and may not be available for complex tasks~\cite{abu2018force}. Impedance control mitigates this limit by controlling the force applied by the robot when it deviates from the desired trajectory and has been introduced to the valve turning task~\cite{abu2018force} and several tools using tasks~\cite{li2018force}.
%  In such tasks, the motion control is not sufficient, and the force control is needed to maintain particular contact and force constraints.
\subsection{Variable Impedance Control}
% Impedance control, which is a position-force control method, establishes a virtual mass-spring-damping relationship between robots and the unknown environment. This method directly controls the force applied by the robot when it deviates from the desired trajectory and allows the safe robot interactions. However, one realistic problem of impedance control is that in many tasks, we may need different impedance gains in different phases of the task. To complete these tasks, Scheduling variable impedance gains is needed to maintain stability or safety for a given kinematic trajectory. 
In many complex tasks, variable impedance skills are needed for robots to interact with the environment. Learning from demonstration approaches is utilized to learn the expert variable impedance skills. In~\cite{peternel2015human}, human experts control a robot's impedance by a hand-held impedance control interface. The expert data is then recorded by dynamical movement primitives (DMP) and learned by the regression method. Researches in~\cite{abu2018force} directly estimated the robot impedance from the human demonstrations and then encoded the robot skill by Gaussian Mixture Regression (GMR) with sensed forces. However, these previous works learned expert variable impedance skills on specific tasks which are difficult to transfer to a different task setting. In our approach, both the expert variable impedance skill and the reward function would be recovered, and new skills can be generated by reoptimizing the learned reward function \RV{with RL}.
% Their approaches can successfully accomplish complex robot manipulation tasks such as valve turning and tight peg in hole. 

In the field of reinforcement learning, Buchli et al.~\cite{buchli2011learning} utilized policy improvement with path integrals ($PI^2$), which is a model-free method to learn the joint space variable impedance skills. However, the joint space impedance they are using limited policy transferability. Rey et al.~\cite{rey2018learning} further improved this method by simultaneously learning trajectories and a state-dependent varying stiffness model. This stiffness skill is now represented in the robot end-effector frame. Martin et al.~\cite{martin2019variable} compared the RL performance of different action spaces in robot manipulation tasks. They showed the variable impedance control in end-effector space (VICES) has an advantage in constrained and contact-rich tasks. However, the results from RL highly depend on the design of the reward function. With our method, the reward function can be recovered from expert demonstrations and generates new policy for different tasks with RL methods.

\subsection{Inverse reinforcement learning}
\RV{Inverse reinforcement learning is one of the learning from demonstrations methods. Unlike the traditional LfD approaches such as BC, where the main idea is to mimic expert actions, the goal of IRL is to infer the expert's cost or reward function from expert demonstrations. Then the optimal policy is obtained by maximizing this reward function using forward RL.}
One commonly used framework for IRL is the maximum entropy IRL ~\cite{ziebart2008maximum}. This framework assumes that the expert trajectories follow a Boltzmann distribution with the cost and updates cost function by maximum likelihood learning. Levine et al.~\cite{levine2012continuous} further improved this framework to the high dimensional and continuous tasks by using the local Laplace approximation of the cost function. However, these methods still require the dynamics model for the cost function update, which is challenging to be obtained in robot manipulation tasks. 

Recently, Finn et al.~\cite{finn2016guided} proposed a sample-based IRL approach to recover cost function in high dimensional state-action spaces. In this method, the agent alternates between optimizing the cost function and optimizing policy, which generates trajectories to minimize the cost. Since the optimizer requires no dynamics model, both the cost function and the policy are updated in a model-free way. Later the authors found that their method agrees with the generative adversarial network (GAN) formulation and introduced the GAN-GCL algorithm in~\cite{finn2016connection}. One practical problem for GAN-GCL is that it evaluates the full trajectory and results in high variance estimates. Adversarial inverse reinforcement learning (AIRL)~\cite{fu2017learning} improved the performance by extending the GAN-GCL algorithm to single state-action pairs and achieved superior results in simulation.

% However, using single state-action pairs also harms the transferability of AIRL. The learned reward function could simply encourage mimicking the expert policy and could fail when task setting changes~\cite{fu2017learning}. 
\RV{Although many IRL algorithms employ entropy regularisation to prevent the simply mimicking the expert policy, there is no previous work focus on the effect of action space selection to the authors' knowledge.} In our method, by introducing variable impedance gain action space, we can find more general representation of the expert policy than using force as action and improve the reward function transfer performance in a new task setting.
\section{PROPOSED APPROACH}\label{PROPOSED APPROACH}

\subsection{Cartesian space impedance control}
Consider the dynamics model of the robot in the Cartesian space:
\begin{equation}
   \label{model:natural model}
    \bm{M}(\bm{x})\bm{\ddot{x}}+\bm{C}(\bm{x},\bm{\dot{x}}) \bm{\dot{x}}+\bm{G}(\bm{\dot{x}})=\bm{J^{-T}}\bm{\tau}-\bm{F}_{ext}
\end{equation}
where $\bm{M}(\bm{x})$ is the mass-inertia matrix, $\bm{C}(\bm{x},\bm{\dot{x}})$ denotes the Coriolis matrix, $\bm{G}(\bm{\dot{x}})$ is the gravity vector, $\bm{\ddot{x}}$, $\bm{\dot{x}}$ and $\bm{x}$ are respectively the Cartesian acceleration, velocity and position of the end-effector, $\bm{J}$ is the Jacobian matrix and $\bm{\tau}, \bm{F}_{ext}$ represents the joint space motor torque input and the external force, respectively. Under the impedance control law, the robot will behave as a mass-spring-damping system, which follows the dynamics equation:
\begin{equation}
   \label{model:impdance model}
    \bm{M_d}(\bm{\ddot{x}-\ddot{x}_d})+\bm{B_d}(\bm{\dot{x}}-\bm{\dot{x_d}}) +\bm{K_d}(\bm{x}-\bm{x_d})=- \bm{F}_{ext}
\end{equation}
where $\bm{M_d},\bm{B_d},\bm{K_d}$ are the desired mass, damping and stiffness matrices. By solving (\ref{model:natural model}), (\ref{model:impdance model}) and setting $\bm{M_d} = \bm{M(x)}$, the impedance control law can be written as:

% \begin{align*}
%     \label{impdeance control law}
%     \bm{\tau} = \bm{J^{T}}\bm{F}\\
%     \bm{F} = \bm{M}\bm{\ddot{x_d}} +\bm{C}\bm{\dot{x}} + \bm{G} +(\bm{I}-\bm{M}\bm{M_d^{-1}})\bm{F}_{ext}\\ + \bm{M}\bm{M_d^{-1}}[\bm{B_d}(\bm{\dot{x}}-\bm{\dot{x_d}}) +\bm{K_d}(\bm{x}-\bm{x_d})]
% \end{align*}
% This is the standard form of impedance control with requires the force feedback. By selecting $\bm{M} = \bm{M_d}$, this control law can be simplified:
\begin{equation}
    \begin{aligned}
    \label{impdeancecontrollaw}
    \bm{\tau} =& \bm{J^{T}}\bm{F}\\
    \bm{F} = \bm{M(x)}\bm{\ddot{x}_d} +&\bm{C(x,\dot{x})}\bm{\dot{x}} + \bm{G(x)}\\  -\bm{B_d}(\bm{\dot{x}}-&\bm{\dot{x_d}}) -\bm{K_d}(\bm{x}-\bm{x_d})
\end{aligned}
\end{equation}

This impedance control law can be further separated into two parts: the feed-forward term $\bm{F_{ff}}$ to cancel the nonlinear robot dynamics and the feedback term $\bm{F_{fb}}$ which tracks the desired trajectory:
\begin{equation}
    \bm{F_{ff}} = \bm{M(x)}\bm{\ddot{x}_d} +\bm{C(x,\dot{x})}\bm{\dot{x}} + \bm{G(x)}
\end{equation}
\begin{equation}
\label{feedback}
\begin{aligned}
    \bm{F_{fb}} &= -\bm{B_d}(\bm{\dot{x}}-\bm{\dot{x_d}}) -\bm{K_d}(\bm{x}-\bm{x_d})\\
    &= -\bm{B_d} \bm{\dot{e}} -\bm{K_d}\bm{e}
\end{aligned}
\end{equation}
where $\bm{e}$ and $\bm{\dot{e}}$ are the tracking error and the tracking velocity. \RV{The stiffness matrix $\bm{K_d}$ and the damping matrix $\bm{B_d}$ are also known as the impedance gain matrices, since they map the tracking error and velocity to the feedback force $\bm{F_{fb}}$. To simplify the notations, we use $\bm{K}$ (stiffness) and $\bm{B}$ (damping) to represent $\bm{K_d}$ and $\bm{B_d}$ in the rest of paper.}
% For dimension reduction, we assume both the stiffness matrix $K_d$ and the damping matrix $B_d$ are diagonal and the variable impedance policy outputs the diagonal component. Furthermore, depending on the task, our policy can output only the stiffness with the fixed damping to further reduce the dimension.

\begin{figure}[H]
     \centering
     \begin{subfigure}[b]{0.47\textwidth}
         \centering
         \includegraphics[scale =0.4]{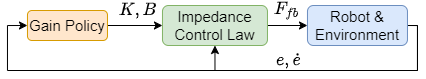}
         \caption{Impedance gain controller}
         \label{fig:gaincontroller}
     \end{subfigure}
     \vfill
     \begin{subfigure}[b]{0.47\textwidth}
         \centering
         \includegraphics[scale=0.4]{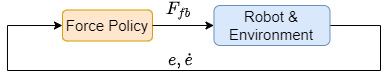}
         \caption{Force controller}
         \label{fig:forcecontroller}
     \end{subfigure}
        \caption{Controller design}
        \label{fig:Controller}
\end{figure}

\subsection{Learning variable impedance skills with AIRL}
Fig.~\ref{fig:Controller} depicts our controller design. In our approach, the observations from the robot and the environment are the tracking error $\bm{e}$ and the tracking velocity $\bm{\dot{e}}$. Our policy takes in the observations and outputs either the impedance gain $\bm{K},\bm{B}$ or the feedback force $\bm{F_{fb}}$, depending on the action space design. The impedance gain controller then calculates the control inputs by (\ref{impdeancecontrollaw}) and controls the robot.

We employ AIRL~\cite{fu2017learning} to learn both the expert policy and the reward function and the training procedure is detailed in Algorithm.~\ref{algo}. In this adversarial training setting, the discriminator which separates the generator trajectories and the expert trajectories is defined as:
\begin{equation}
    D_\theta(o,a) = \frac{exp(r_\theta(o,a))}{exp(r_\theta(o,a))+\pi(a|o)}
\end{equation}
where $r_\theta(o,a)$ is the reward function we want to learn and $\pi(a|o)$ is the probability of taking action $a$ at observation $o$ under current policy. The discriminator is updated to minimize this loss~\cite{fu2017learning}:
\begin{equation}
    L(\theta) =\sum_{t=0}^{T} -\mathbb{E}_{D}[logD_\theta(o_t,a_t)]-\mathbb{E}_{\pi_t}[1- logD_\theta(o_t,a_t)]
\label{IRL-loss}
\end{equation}
% This updating process is essentially the reward function updating. 
% In AIRL, the discriminator is updated to minimize this loss：

The generator is the variable impedance policy. During the training, the policy is updated to maximize the trajectory reward, which is evaluated by the reward function. In our approach, we use TRPO~\cite{schulman2015trust}\RV{, which is a policy gradient based RL method, for the policy update.}

Since the environment dynamics are unknown, we applied RL to reoptimize a new policy in a different task setting to test the learned reward function's performance. In the RL process, the policy update is the same as the IRL but with the fixed learned reward function.

\begin{algorithm}
\SetAlgoLined

 Collect expert trajectories $\tau_i^E$

 Initialize impedance gain policy $\pi$ and reward function $r$ with random weights
 
 \For{iteration k = 1 to K}{
 Collect trajectories $\tau_i$ under policy $\pi$
 
 Update the reward function by minimizing (\ref{IRL-loss})
 
 Evaluate trajectories $\tau_i$ with the reward function
 
 Update policy $\pi$ to maximize the reward by TRPO
 }
 \caption{Learn variable impedance skills}
 \label{algo}
\end{algorithm}
% In the first comparison, we compare the AIRL policy performance and the cost function transferability in these two action spaces. For the second comparison, we compare the performance of the learned policy from AIRL and the reoptimized policy using learned cost function with the BC policy. Since the force policy doesn't contain the feedback law, their performance in the new task setting should be lower than the variable impedance policy. The BC policy should have similar performance with the expert in the learning scenario but can fail in the testing scenario due to unseen trajectory during the training. The comparison results are shown in section \ref{EXPERIMENTS}.
\begin{figure}
    \centering
     \begin{subfigure}[b]{0.2\textwidth}
         \centering
         \includegraphics[scale =0.18]{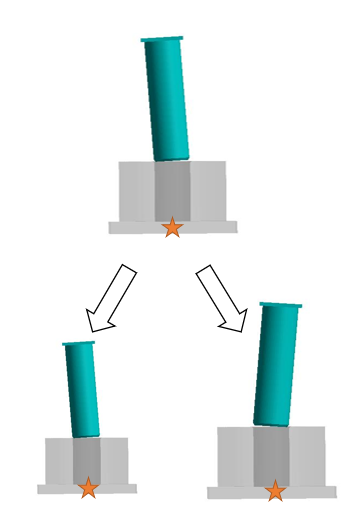}
         \caption{The Peg-in-Hole task}
        %  \label{fig:pih_tilt}
     \end{subfigure}
     \hfill
     \begin{subfigure}[b]{0.26\textwidth}
         \centering
         \includegraphics[scale=0.23]{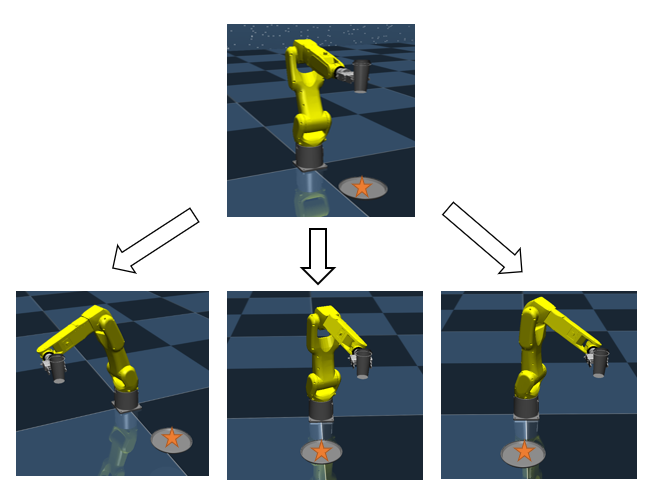}
         \caption{The Cup-on-Plate task}
        %  \label{fig:pih_size}
     \end{subfigure}
        \caption{Training and testing scenarios for a) the Peg-in-Hole task and b) the Cup-on-Plate task, orange stars indicate the goal points}
        \label{fig:envs}
\end{figure}
% \begin{figure*}
%     \centering
%     \includegraphics[scale=0.3]{figures/envs2.png} 
%     \caption{Training and testing scenarios for the Peg-in-Hole task and Cup-on-Plate task: a) the Peg-in-Hole task, b) the Cup-on-Plate task. The orange stars indicate the goal point}
%     \label{fig:envs}
% \end{figure*}
\section{EXPERIMENTS}\label{EXPERIMENTS}
To evaluate our proposed approach, we conducted experiments in several robotic tasks, both in simulation and on a real robot. In the simulation, we designed two variable impedance control tasks: Peg-in-Hole task and Cup-on-Plate task. For the experiment validation, we collected human expert data for the Cup-on-Plate task by kinesthetic teaching and tested learned policies on the real robot. 

To validate our approach's generalizability, we compare the performance of our approach with four baseline methods: 1) force-based AIRL, 2) gain-based BC, 3) force-based BC and 4) constant gain.  We hypothesize that using the impedance gain as action space improves the transferability to the testing scenarios, either via AIRL or BC, since the impedance control law remains effective when task setting changes. Furthermore, by recovering the expert reward function, leaning gain in the impedance control via AIRL is a more general approach than BC.

\subsection{Tasks in simulator}
\subsubsection{Task setups}

\hfill

As depicted in Fig.~\ref{fig:envs}, two robotic tasks, Peg-in-Hole and Cup-on-Plate, are conducted in the Mujoco \cite{todorov2012mujoco} simulation to evaluate the proposed approach. 
% In the Peg-in-Hole task, the agent's goal is to drive a round peg into a round hole in the shortest time. For the Cup-on-Plate task, the robot's goal is to place a cup on a plate in a fast and steady manner. 
\RV{In the Peg-in-Hole task, the diameters of the peg and hole are $25.37$ mm and $25.40$ mm, respectively. For the Cup-on-Plate task, FANUC LR Mate 200iD/7L industrial robot is included into simulation and the robot's goal is to place a cup on a plate in a fast and steady manner. In two simulation environments, model parameters such as the mass-inertia matrix $\bm{M(x)}$, the Coriolis matrix $\bm{C(x,\dot{x})}$ and the gravity vector $\bm{G(x)}$ are calculated by Mujoco automatically using the simulation model.}
% To achieve this goal, the robot needs to choose appropriate control gain according to the current observation. 
\subsubsection{Observation spaces}

\hfill

We use the tracking error $e$ and tracking velocity $\dot{e}$ together as observations for two tasks. The goal points are shown in Fig.~\ref{fig:envs}, and the end-effectors are located on the center of mass (COM) of the peg for the Peg-in-Hole task and on the cup for the Cup-on-Plate task.

\RV{Moreover, since a single pair of $e$ and $\dot{e}$ doesn't provide acceleration information and may not fully represent the system dynamics. We also employ an augmented observation which contains a history of $e$ and $\dot{e}$ from the last five time steps for evaluation. In Table.~\ref{tab:PIH_tilt},\ref{tab:PIH_size} and \ref{tab:COP_PS}, evaluations on the augmented observation space are marked with "His" which denotes using a history of $e$ and $\dot{e}$.}

% We also employ an augmented observation $O_{aug~ t} = \{O_{t-4},...,O_{t}\}$ to evaluate our approach, which contains a history of last five $e$ and $\dot{e}$.

\subsubsection{Action spaces}

\hfill

There are two options for the action space selection, the force action, and the impedance gain action. For the force action case, our policy outputs the Cartesian space force exerted by the robot. For the impedance gain action space, our policy outputs impedance gains and the control input are obtained by equation (\ref{feedback}).

\RV{To reduce the dimension of the gain action space, we suppose that the stiffness matrix $\bm{K}$ and damping matrix $\bm{B}$ are diagonal. Therefore, our policy outputs the diagonal elements of two matrices rather than the full matrix. Furthermore, by enforcing the diagonal elements to be positive, we can ensure the stiffness matrix $\bm{K}$ and damping matrix $\bm{B}$ are positive definite. To extend our approach to the full matrix case, Cholesky decomposition can be utilized to guarantee $\bm{K}, \bm{B} >0$.}

Different impedance gain outputs are utilized for two tasks according to task requirements. In the Peg-in-Hole task, the peg velocity is small, and the stiffness plays a main role in the insertion. Thus, the policy outputs a 6-dimensional stiffness, $[K_1, K_2, K_3, K_4, K_5, K_6]$ with fixed damping term, \RV{where $K_i~(i=1,2,3)$ denote the stiffness for the position error and $K_i~(i=4,5,6)$ denote the stiffness for the orientation error}. For the Cup-on-Plate task, the tracking velocity is large, and the damping term affects the performance. Therefore, \RV{the output of our policy is now $[K_1,...,K_6,d]$, which contains an extra damping factor $d$. The stiffness and damping matrices can then be obtained by:}
% the gain matrices are:
\begin{align*}
    K = diag(K_1, K_2, K_3, K_4, K_5, K_6) \quad B = d\sqrt{K}
\end{align*}
%where $d$ is the damping factor. 
We use a 1-dimensional damping factor rather than another 6-dimensional damping for dimension reduction.
% We use different impedance gain outputs for two tasks. In the Peg-in-Hole task, the policy outputs a 6-dimensional stiffness, $[K_1, K_2, K_3, K_4, K_5, K_6]$, and the damping term is fixed. For the Cup-on-Plate task, we have an extra damping factor term and the gain matrices are:
% \begin{align*}
%     K = diag(K_1, K_2, K_3, K_4, K_5, K_6) \quad D = \sqrt{d}K
% \end{align*}
% where $d$ is the damping factor.

\subsubsection{Generating expert data}

\hfill

Fifty trajectories are collected for both tasks by two designed variable impedance controllers to generate expert data. In the Peg-in-Hole task, the expert is a fixed tip stiffness controller. The stiffness of the peg tip can be transferred to the COM frame by:
\begin{equation}
    K_{COM} = J_{tip}^T \ K_{tip} \ J_{tip}
\end{equation}
where \RV{$K_{tip}$ denotes the stiffness matrix of the peg tip and }$J_{tip}$ is the Jacobian matrix between COM and the tip, which depends on the peg's configuration. Thus, this fixed tip stiffness is a variable stiffness on the COM. The uncorrelated stiffness $diag(K_1, K_2, K_3, K_4, K_5, K_6)$ is solved by:
\begin{equation}
    diag(K_1, K_2, K_3, K_4, K_5, K_6) \ e = K_{COM}e
\end{equation}
where e is the tracking error.

In the Cup-on-Plate task, the variable impedance controller contains three phases:
\begin{align*}
    controller \ phases=
     \begin{cases}
      accelerating, \ &\text{if } e_{pos}>e_1\\
      switching, \ &\text{if } e_1>e_{pos}>e_2 \\
      reaching, \ &\text{if } e_2>e_{pos} \\
     \end{cases}\\
\end{align*}
where $e_{pos} = \sqrt{e_x^2+e_y^2+e_z^2}$ indicates the Cartesian space position tracking error and $e_1,e_2$ are two gain changing points which are $0.4$~m and $0.2$~m. As shown in Fig.~\ref{fig:COP_com_learning}, our designed expert control law chooses the largest gain to accelerate in the accelerating phase and generally switching to the smaller gain in the switching phase. In the reaching phase, the robot approaches the plate with minimum speed to guarantee safety.
\subsubsection{Performance score}

\hfill

In the experiment, we designed performance functions to evaluate different policies. In the Peg-in-Hole task, the performance function outputs a constant penalty if the peg is not aligned with the hole. Otherwise, it penalizes the $z$ direction error, which is the vertical distance to the target point. 

% Therefore, if one policy accomplishes the Peg-in-Hole task in a short time, it will have a high performance function score.

The designed performance function in the Cup-on-Plate task only penalizes the tracking error $e$ when in the accelerating phase and the tracking velocity $\dot{e}$ in the reaching phase. If in the switching phase, the the performance function penalizes both $e$ and $\dot{e}$. 

% Thus, the optimal policy should minimize the tracking error $e$ in the first phase and then gradually change to the smallest stiffness and largest damping to slowly place the cup on the plate.
\RV{
\subsubsection{Training details}

\hfill

For the network architectures, our reward function is a 2-layer neural network with 32 units. For the policy, we use a 2-layer gaussian policy with 32 units when using a single pair of $e$ and $\dot{e}$ as observation. The units number increases to 128 for the augmented observation space. The activation function is Tanh for the policy and Relu for the reward function.

For the Learning hyper-parameters, the batch size and trajectory length of TRPO are 8000 and 200, respectively in the Peg-in-Hole task. For the Cup-on-Plate task, we use a batch size of 10000 and trajectory length of 500.
}

\subsubsection{Evaluations}

\hfill

In evaluation, we set two tests to compare different approaches, which are the imitation learning performance in the training scenario and the transfer learning performance in the testing scenarios.
% In the evaluation, we first train policies and reward functions of four approaches in the training scenario. After training, their imitation performances will be tested. We then compare the generalizability of four approaches in the testing scenarios with different task settings. 
% In the experiment, four policies (gain-based AIRL, force-based AIRL, gain-based BC and force-based BC) are trained in the training scenario and then transferred to the testing scenarios with different task settings. For evaluation, we test their abilities of imitating expert policy in the training and the transfer learning results in the testing.
\hfill

\begin{figure}
    \centering
     \begin{subfigure}[b]{0.4\textwidth}
         \centering
         \includegraphics[scale =0.35]{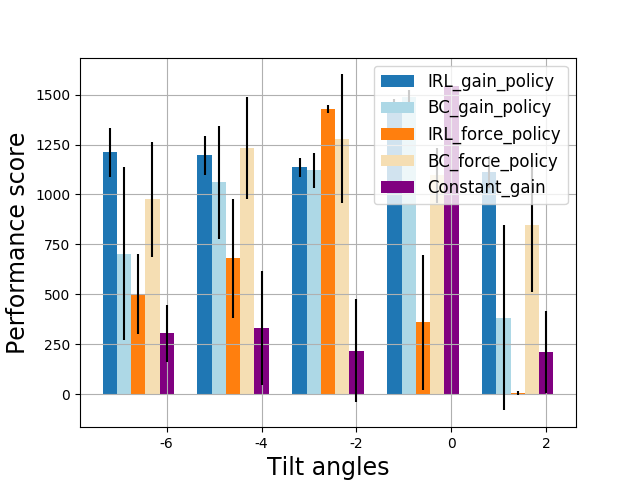}
         \caption{Testing scenarios with different tilt angles}
         \label{fig:pih_tilt}
     \end{subfigure}
     \vfill
     \begin{subfigure}[b]{0.4\textwidth}
         \centering
         \includegraphics[scale=0.35]{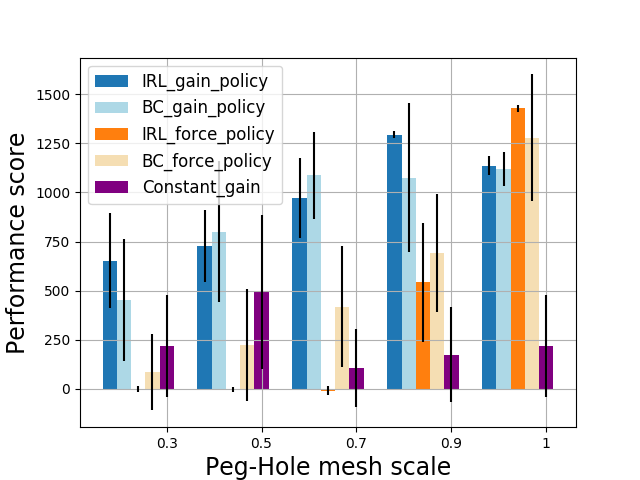}
         \caption{Testing scenarios with different mesh scales}
         \label{fig:pih_size}
     \end{subfigure}
        \caption{Transfer performance on the  Peg-in-Hole task}
        \label{fig:pih_testing}
\end{figure}
\textbf{Peg-in-Hole task:} 
We first evaluate the imitation learning results of four policies in the training scenario. The mean performance score and the mean success rate of three runs are given in the rightmost column of  Fig.~\ref{fig:pih_size} and marked with (T) in Table.~\ref{tab:PIH_tilt} and Table.~\ref{tab:PIH_size}. We observe that all policies achieve high success rates in accomplishing the Peg-in-Hole task and their performance is similar, which approves these polices successfully imitate the expert in the training.

We then transfer the learned reward functions obtained by AIRL to generate new policies in the testing scenario. Since Peg-in-Hole is a difficult task, the initial policy influences the final reoptimizing results. In one setting, we select the three best policies from five random runs to evaluate the transfer performance to mitigate this effect. For BC, we directly transfer the policy obtained in training to the testing scenarios.

\begin{table}[H]
    \centering
    \caption{Peg-in-Hole success rate in different tilt angles}
    \label{tab:PIH_tilt}
  \begin{tabular}{cccccc}
    \cline{1-6}
      Tilt angles& $-6^\circ$& $-4^\circ$& $-2^\circ$(T) &$0^\circ$&$2^\circ$ \\
    \cline{1-6}
    % Expert & $100\%$ &$100\%$ \\
    Gain-AIRL & $\bm{91.7\%}$ & $\bm{100\%}$ & $98.3\%$ & $\bm{100\%}$ & $\bm{100\%}$ \\
    
    Gain-BC & $56.0\%$ & $100\%$ & $96.7\%$ & $100\%$ & $46.7\%$ \\
    
    Force-AIRL & $53.3\%$ & $71.7\%$ & $\bm{100\%}$ & $56.7\%$ & $0\%$  \\
    
    Force-BC & $80.0\%$ & $95.0\%$ & $93.3\%$ & $100\%$ & $75.0\%$ \\
    
    Constant-Gain & $6.7\%$ & $10.0\%$ & $36.7\%$ & $100\%$ & $31.7\%$\\
    \cline{1-6}
    Gain-AIRL-His & $\bm{100\%}$ & $\bm{91.7}\%$ & $\bm{100\%}$ & $\bm{100\%}$ & $\bm{93.3\%}$\\
    Gain-BC-His & $63.3\%$ & $86.7\%$ & $90.0\%$ & $100\%$ & $55.0\%$\\
    Force-AIRL-His & $20.0\%$ & $0\%$ & $75.0\%$ & $21.6\%$ & $31.6\%$\\
    Force-BC-His & $86.7\%$ & $86.7\%$ & $93.3\%$ & $83.3\%$ & $83.3\%$\\
    
    \cline{1-6}
      \end{tabular}
    \end{table}
\begin{table}[H]
    \centering
    \caption{Peg-in-Hole success rate in different mesh scales}
    \label{tab:PIH_size}
  \begin{tabular}{cccccc}
    \cline{1-6}
      Mesh scale& 0.3& 0.5& 0.7 &0.9&1.0(T)\\
    \cline{1-6}
    % Expert & $100\%$ &$100\%$ \\
     Gain-AIRL & $\bm{91.7\%}$ & $\bm{90.0\%}$ & $\bm{95.0\%}$ & $\bm{100\%}$ & $98.3\%$ \\
    
    Gain-BC & $70.0\%$ & $80.0\%$ & $93.3\%$ & $91.7\%$ & $96.7\%$ \\
    
    Force-AIRL & $0\%$ & $0\%$ & $0\%$ & $91.6\%$ & $\bm{100\%}$  \\
    
    Force-BC & $16.7\%$ & $40.0\%$ & $75.0\%$ & $90.0\%$ & $93.3\%$ \\
    Constant-Gain & $28.3\%$ & $43.3\%$ & $11.7\%$ & $23.3\%$ & $36.7\%$\\
    \cline{1-6}
    Gain-AIRL-His & $\bm{86.7\%}$ & $\bm{90.0}\%$ & $\bm{91.7\%}$ & $\bm{100\%}$ & $\bm{100\%}$\\
    Gain-BC-His & $33.3\%$ & $66.7\%$ & $83.3\%$ & $83.3\%$ & $90.0\%$\\
    Force-AIRL-His & $0\%$ & $21.7\%$ & $0\%$ & $26.7\%$ & $75.0\%$\\
    Force-BC-His & $6.7\%$ & $50.0\%$ & $63.3\%$ & $76.7\%$ & $93.3\%$\\
    
    \cline{1-6}
      \end{tabular}
    \end{table}

\begin{figure*}[htp]
    \centering
     \begin{subfigure}[b]{0.49\textwidth}
         \centering
         \includegraphics[scale =0.4]{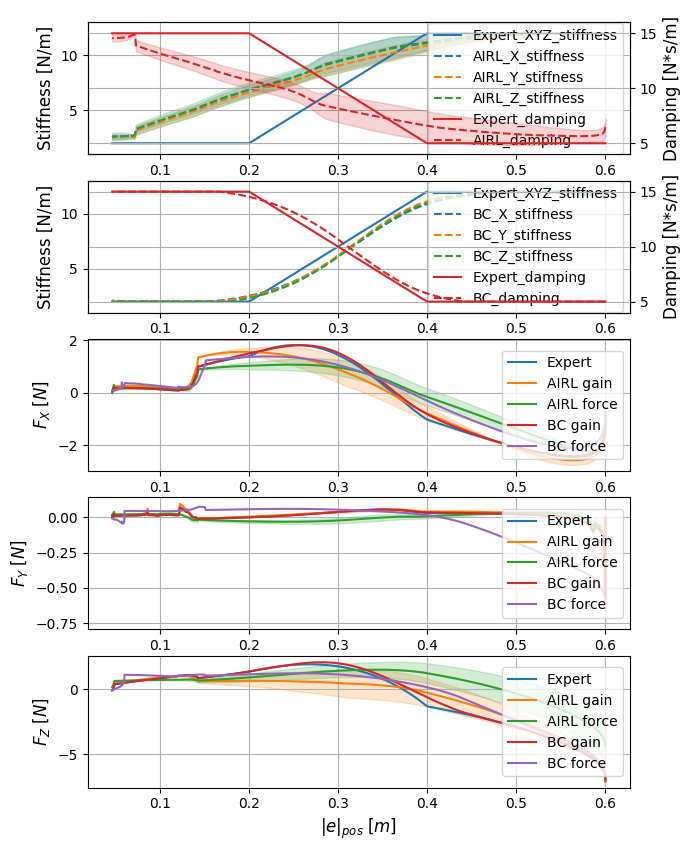}
         \caption{Policy comparisons in the training scenario}
         \label{fig:COP_com_learning}
     \end{subfigure}
     \hfill
     \begin{subfigure}[b]{0.49\textwidth}
         \centering
         \includegraphics[scale=0.4]{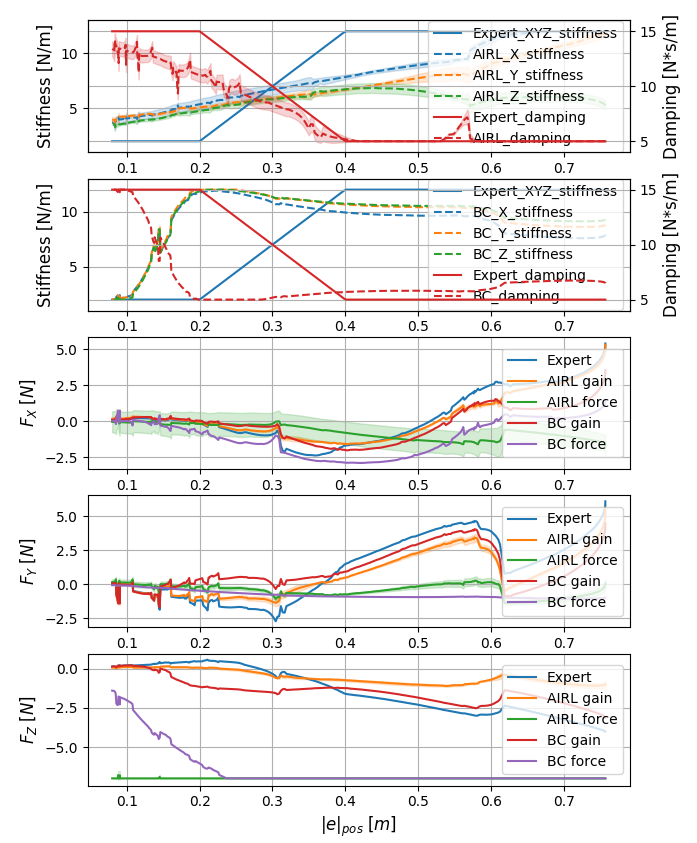}
         \caption{Policy comparisons in the testing scenario}
         \label{fig:COP_com_testing}
     \end{subfigure}
        \caption{Policy comparisons for Cup-on-Plate task: two top figures of a) and b) compare the policy outputs of gain-based AIRL and BC with the expert gain policy; three bottom figures of a) and b) show the force applied by policies in $x,y,z$ directions}
        \label{fig:COP_COM}
\end{figure*}

In the first testing scenario, we change the tilt angles of the peg. Results are depicted in Fig.~\ref{fig:pih_tilt} and Table.~\ref{tab:PIH_tilt}. We find that gain-based AIRL achieves the highest success rate and outperforms all the other policies. Two BC policies achieve a high success rate when the tilt angle is close to $-2^\circ$, but fails to generalize to tilt angles $-6^\circ$ and $2^\circ$, where trajectories deviate from expert demonstrations. The force-based AIRL has zero success rate when the tilt angles are $2^\circ$. In this case, the force policy imitates the expert action in the training, pushing the peg in the wrong direction from the hole and falls. However, since the sign of the tracking error $e$ changes, the feedback force generated by the impedance control law also reverses the direction. Thus the gain policy is still valid when the tilt angle changes and avoids getting stuck in the middle. \RV{For the constant gain baseline, it achieves 100\% success rate when the tilt angle is $0^\circ$, which is the simplest scenario. However, since it doesn't utilize the variable impedance control gain, its success rate is much lower than our proposed approach when increasing the tilt angle.}

For the second testing scenario, we use smaller mesh scales of the peg and hole. Fig.~\ref{fig:pih_size} depicts the performance scores results. We notice that using gain as action improves the transferability, either for AIRL or BC. The reason is that, \RV{as shown in Fig.~\ref{fig:Controller}, the feedback force of the gain policy is generated by the impedance control law, which is still effective when task setting changes. However, the force policy doesn't have this advantage and could be infeasible in the new dynamics}. The performance score difference is negligible for gain-based AIRL and BC. However, as shown in Table.~\ref{tab:PIH_size}, the gain-based AIRL achieves higher successful rate when the mesh scale is small. 

% The reason is that, with the smaller peg and hole, the expert force policy is not feasible in the new dynamics. In contrast, the gain policy still holds in this case.

\RV{We also tested our approach with all the baselines on the augmented observation space, which contains the last five tracking errors and velocities. As shown in Table.~\ref{tab:PIH_tilt} and Table.~\ref{tab:PIH_size}, our approach still achieves the highest success rate and has similar performance in comparison with the original observation space. However, since the dimension of the augmented observation space is much larger, the force-based AIRL policy is more difficult to train and has even worse generalization results.}

\begin{table}
    \centering
    \caption{Cup-on-Plate relative performance difference}
    \label{tab:COP_PS}
  \begin{tabular}{cccccc}
    \cline{1-6}
      & Training & T1& T2&T3&T4 \\
    \cline{1-6}
    Gain-AIRL & $\bm{0.01}$ & $\bm{0.09}$ & $\bm{0.01}$ &$0.19$& $\bm{0.24}$\\
    
    Gain-BC & $0.02$  & $0.17$& $0.10$ &$0.25$& $0.36$\\
    
    Force-AIRL & $0.1$ & $0.55$ & $0.58$ &$0.32$& $0.26$\\
    
    Force-BC & $0.05$ & $0.40$& $0.20$ &$0.42$& $0.58$\\
    
    Constant-Gain & $0.14$ & $0.18$ & $0.18$ &$\bm{0.17}$& $0.39$\\
    \cline{1-6}
    Gain-AIRL-His & $0.04$ & $\bm{0.11}$ & $\bm{0.01}$ &$\bm{0.12}$& $\bm{0.38}$\\
    
    Gain-BC-His & $0.04$  & $0.12$& $0.12$ &$0.15$& $0.53$\\
    
    Force-AIRL-His & $\bm{0.02}$ & $0.69$ & $0.53$ &$0.51$& $0.54$\\
    
    Force-BC-His & $0.06$ & $0.25$ & $0.29$ &$0.36$& $0.65$\\

    % IRL gain & $3.1$ & $-60.3$ & $-5.6$ &$-97.7$& $-198.0$\\
    
    % BC gain & $-7.0$  & $-101.9$& $-49.3$ &$-122.7$& $-293,8$\\
    
    % IRL force & $-33.7$ & $-377.8$ & $-278.2$ &$-160.6$& $-204.6$\\
    
    % BC force & $-16.9$ & $-242.5$& $-97.1$ &$-209.6$& $-464.7$\\
    % Expert & $-343.2$ &$-617.6$ & $-476.6$ &$-503.3$& $-507.7$ \\
    
    % IRL gain & $-340.1$ & $-677.9$ & $-482.2$ &$-601.0$& $-705.7$\\
    
    % % RL gain & $\pi$ & 440 \\
    
    % BC gain & $-350.2$  & $-719.5$& $-525.9$ &$-626.0$& $-801.5$\\
    
    % IRL force & $-376.9$ & $-995.4$ & $-754.8$ &$-663.9$& $-712.3$\\
    
    % BC force & $-360.1$ & $-860.1$& $-573.7$ &$-712.9$& $-972.4$\\
    \cline{1-6}
      \end{tabular}
    \end{table}

% similar performance in compare with the original observation space and outperforms four baselines. However, the  
\begin{figure*}
    \centering
    \includegraphics[scale=0.35]{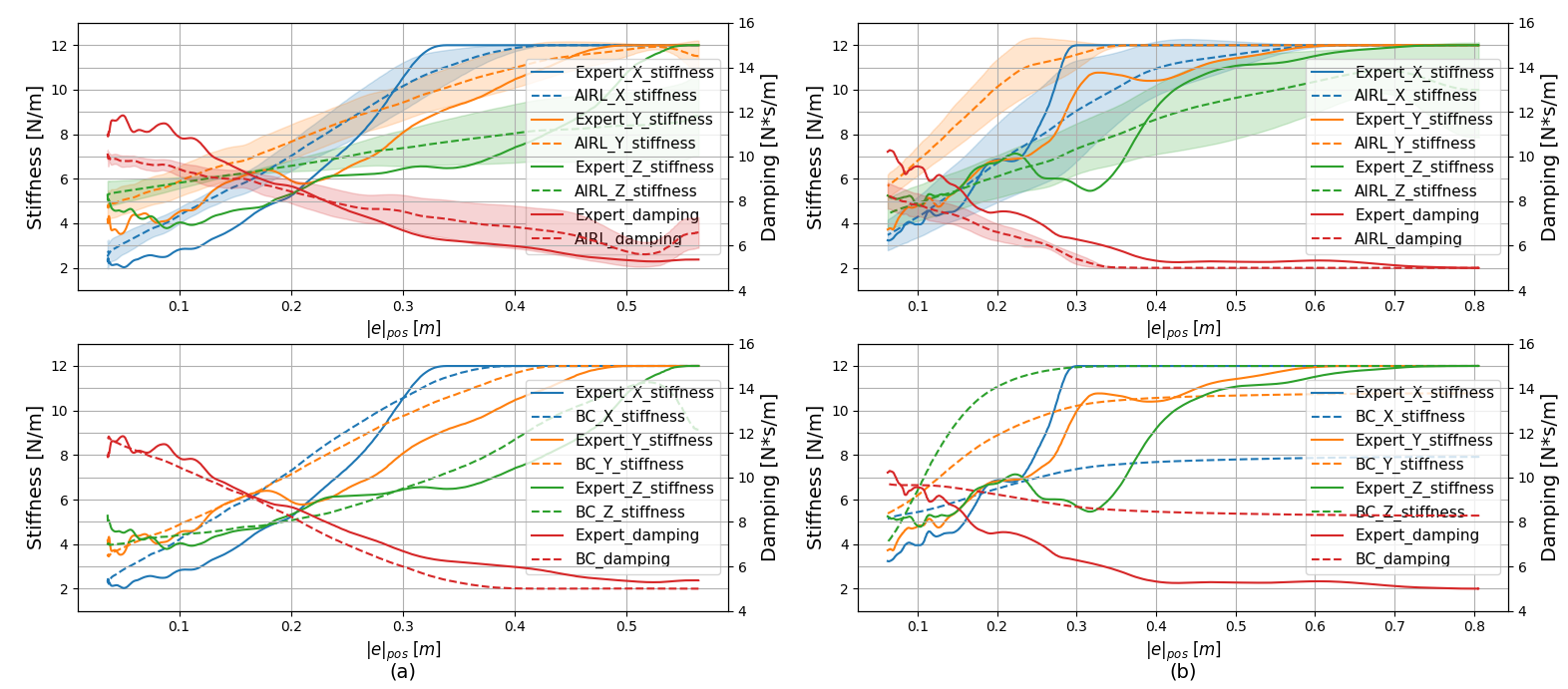}
    \caption{Policy comparisons for the real robot: a) gain-based AIRL and BC policies in the training, b) gain-based AIRL and BC policies in the testing}
    \label{fig:real_world}
\end{figure*}
    
\textbf{Cup-on-Plate task:} 
The first column of Table.~\ref{tab:COP_PS} shows the relative performance difference with the expert in training. This difference is a normalized performance score difference with the expert, and a larger value indicates a larger performance difference. In the training scenario, all four policies accomplish the Cup-on-Plate task and achieve a similar performance score compared to the expert. We also compare the gains and forces applied by learned policies, and the results are depicted in Fig.~\ref{fig:COP_com_testing}. We observe that both the gain-based AIRL and the force-based AIRL recover the most features of the expert. \RV{Since the objective of BC encourages the policy to directly mimic the expert action, two BC baselines achieve the best performance in imitating expert gain and force policy. Two AIRL policies are obtained by maximizing the learned reward function, therefore fall behind with two BC baselines.} 
% However, two BC baselines achieve the best performance in imitating expert gain and force policy. 

In the testing, the robot end-effector is initialized in four different positions. The performance scores in comparison with the expert are given in Table.~\ref{tab:COP_PS}, and four initial testing points are named from T1 to T4. As a result, gain-based AIRL outperforms other policies in \RV{T1, T2, T4 and only falls behind the constant gain baseline in T3.}
% four testing scenarios.

We explore the underlying reason for this difference by comparing the gains and forces. Fig.~\ref{fig:COP_com_testing} depicts the policy comparison results for T1. In comparison, the gain-based AIRL recovers the expert gain changing trend and the expert force in $x,y$ directions. However, it chooses small stiffness in $z$ direction in the beginning and applies a much smaller force than the expert. This result may relate to the changes of robot dynamics in the new initial point. \RV{Since the mass matrix $M(x)$ in (\ref{impdeancecontrollaw}) is related to the current robot configuration, the Cartesian space dynamics is coupled.} We observe that, under the new setting, the force applied in $y$ direction will also result in an acceleration in $z$ direction, while this effect is negligible in the training scenario. \RV{This dynamics change could influence the way to maximize the reward function and therefore influences the learned policy.}

% This difference is related to the Cartesian space robot dynamics and influences the learning results. 

For gain-based BC, it learns to set a small stiffness and large damping when the error is small. However, the gain switching happens too late, and the performance score is penalized a lot for the final velocity. As depicted in Fig.~\ref{fig:COP_com_testing}, the force-based AIRL applies force in the wrong direction in $x,y$ axes, and fails in the testing scenario. The reason is that the learned reward function for the force action encourages imitation of the expert force actions, which cannot drive the cup to the plate with the new initial point. 

% On the contrary, the gain action's performance is guaranteed by the feedback control law and thus is better in the testing scenario.

\RV{The evaluation results using the augmented observation space are shown in Table.~\ref{tab:COP_PS}. The augmented observation space doesn't improve the performance either for the gain action or for the force action in the training and testing and our approaches still outperforms all the baselines.}
% \begin{table}[tp]  
%     \centering
%     \caption{Cup-on-Plate average performance score}
%     \label{tab:COP_PS}
%   \begin{tabular}{ccc}
%     \cline{1-3}
%       & learning performance score & testing performance score \\
%     \cline{1-3}
%     Expert & $-343$ &$-622$ \\
    
%     IRL gain & $-340.18$ & $-677.99$ \\
    
%     % RL gain & $\pi$ & 440 \\
    
%     BC gain & $-350.2$  & $-719.5$ \\
    
%     IRL force & $-376.97$ & $-1010.42$ \\
    
%     BC force & $-360.1$ & $-860.1$\\
%     \cline{1-3}
%       \end{tabular}
%     \end{table}
\begin{table}
    \centering
    \caption{Average trajectory deviation on the real robot}
    \label{tab:REAL_WORLD_TABLE}
  \begin{tabular}{ccccc}
    \cline{1-5}
      & Gain-AIRL & Gain-BC & Force-AIRL & Force-BC\\
    \cline{1-5}
    
    Training & $12.6$ mm & $9.5$ mm & N/A & $35.2$ mm\\
    
    Testing& $13.4$ mm & $48.5$ mm & N/A & N/A\\
    \cline{1-5}
      \end{tabular}
    \end{table}

\begin{table}
    \centering
    \caption{Final deviation on the real robot}
    \label{tab:REAL_WORLD_TABLE_final}
  \begin{tabular}{ccccc}
    \cline{1-5}
      & Gain-AIRL & Gain-BC & Force-AIRL & Force-BC\\
    \cline{1-5}
    
    Training & $12.1$ mm & $12.0$ mm & N/A & $70.9$ mm\\
    
    Testing& $10.8$ mm & $17.0$ mm & N/A & N/A\\
    \cline{1-5}
      \end{tabular}
    \end{table}

\subsection{Experiments on the real robot}

We also evaluate our method on the real-world Cup-on-Plate task. In the real-world experiment, expert data is collected by a human expert on the real robot. Then the learned policies are transferred to the real robot for the performance evaluation. 
% Since the simulation has shown that the impedance gain action space has more generalizable results than the force action space, we only use the gain as action in the real-world experiment.

\subsubsection{Task setup}

\hfill

The real-world experimental setup consists of the host and target computer, the F/T sensor, and the FANUC LR Mate 200iD robot. We programmed the Cartesian variable impedance control algorithm on the host PC and it controls the real robot system which is connected to the target PC via Simulink Real-Time. \RV{\RV{The model parameters of the real robot such as the mass-inertia matrix $\bm{M(x)}$, the Coriolis matrix $\bm{C(x,\dot{x})}$ and the gravity vector $\bm{G(x)}$ are obtained by the Euler-Lagrange method.}}

% The real-world robot shares the same model parameters and the control gain range as the previous simulation. We bypassed the built-in controller of the robot and control it with a Cartesian variable impedance control algorithm programmed on Simulink Real-Time of the host PC.

% The Cartesian variable impedance control algorithm is programmed on the host computer and applied to the robot via Simulink Real-Time.The real-world robot shares the same impedance control law and the control gain range as the previous simulation.

\begin{figure}
    \centering
     \begin{subfigure}[b]{0.23\textwidth}
         \centering
         \includegraphics[scale =0.26]{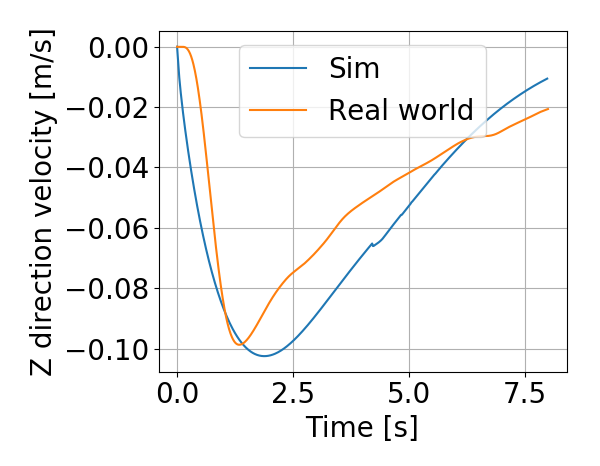}
         \caption{}
         \label{fig:sim2real_gain}
     \end{subfigure}
     \hfill
     \begin{subfigure}[b]{0.23\textwidth}
         \centering
         \includegraphics[scale=0.26]{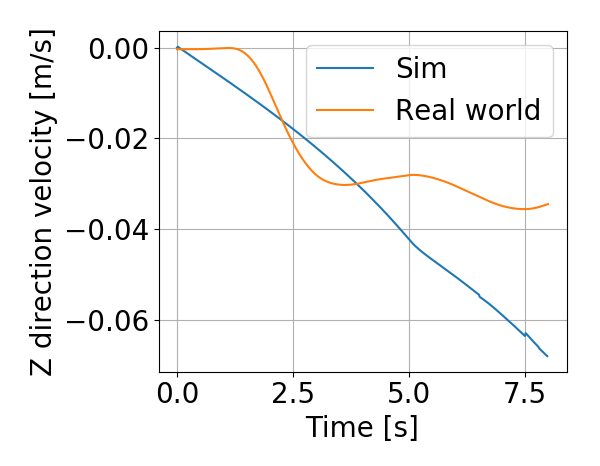}
         \caption{}
         \label{fig:sim2real_force}
     \end{subfigure}
        \caption{Velocity response of the robot in simulation and real-world to same a) gain action and b) force action}
        \label{fig:sim2real}
\end{figure}
% Velocity response of the robot in simulation and real-world to same a) gain action and b) force action Velocity response of the robot in simulation and real-world of a) tracking with same gain b) applying same force
\subsubsection{Collecting human expert data}

\hfill

In the data collection process, the human expert applies force and torque on the end-effector to place cup on the plate. This 6-dimensional Cartesian space force and torque are measured by the F/T sensor and the control input is then calculated with (\ref{impdeancecontrollaw}). We record both the tracking state $[e,\dot{e}]$ and the human expert force together as the human expert data. The human expert gain is estimated in the data processing.
We collect thirty expert trajectories with the same initial point as the training scenario in the simulation. 
% , which is mounted between the robot's sixth joint and the gripper
\subsubsection{Gain estimation using sliding window method}

\hfill

To recover the expert gain policy, we apply a similar method as~\cite{abu2018force}, which uses a short sliding window to estimate stiffness and damping from the force. Each time window contains ten state-force pairs, and expert gain is estimated by solving (\ref{feedback}) with Least Square. 

% Our method is a simplified version without symmetric and positive definite (SPD) approximation. Since we have already assumed the full stiffness and damping matrices are diagonal, and the expert force follows the negative feedback control law, the recovered stiffness and damping matrices are always SPD. 

\subsubsection{Evaluation}

\hfill

With real-world human expert data, we learned both a policy and a reward function in the simulation environment by using AIRL. Training settings other than expert data remain the same as the simulation Cup-on-Plate task.

% Unlike the deterministic expert policy in the previous simulation tasks, the real-world human expert policy is stochastic and noisy. Thus, the performance function in the simulation task is not sufficient to describe the expert policy and we directly compare the impedance gains generated by the learned policies with the expert for evaluation.

\RV{We first evaluate the performance of two gain action approaches,which are gain-based AIRL and BC.} Fig.~\ref{fig:real_world}a depicts the real-world expert gains and the recovered gain policies. Although the expert policy is noisy and the Sim-to-Real gap exists, both the gain-based AIRL and BC successfully recover the expert gains. We then collect three trajectories with the learned policy on the real robot. 
% and record the average trajectory deviation with the expert. 
As shown in Table.~\ref{tab:REAL_WORLD_TABLE} and \RV{Table.~\ref{tab:REAL_WORLD_TABLE_final}}, average deviations with the expert of the gain-based AIRL and BC are close to 10 mm \RV{and the final deviations from the target point are both 12 mm}. 

We also evaluate \RV{two gain action approaches} in the testing scenario with a different initial position. As shown in Fig.~\ref{fig:real_world}b, the BC policy tends to apply constant damping and $x$ direction stiffness in the testing scenario and causes lagging in comparison with the expert trajectory. However, the gain-based AIRL successfully recovers the expert variable impedance policy. Thus, in the testing, the average deviation from the expert and the final deviation are 13.4 mm and 10.8 mm, respectively, which are much smaller than the gain-based BC and in the same range as in the training scenario. 

\RV{For the force-based AIRL, it cannot finish the Cup-on-Plate task and finally drives the robot out of the feasible workspace, both in the training and testing. Its failure is caused by the large sim-to-real gap on the force action space. As shown is Fig.~\ref{fig:sim2real}, the velocity responses of the same force action in simulation and real-robot is too different which makes the learned policy in simulation hard to directly transfer to the real-world. However, for the gain policy, the feedback impedance control gain prevents the trajectory from deviating too much from the target point and therefore result in a much smaller sim-to-real gap for policy transfer. For the force-based BC, it can reach the target point in the training but with a much larger final deviation than two gain based methods. Moreover, it also fails in the testing scenario.}

\section{CONCLUSIONS}\label{conclusion}
In this paper, we introduced an IRL based approach to recover variable impedance policies and reward functions from expert demonstrations. While the learned policy can be utilized in the original task, new impedance policies can be generated by optimizing the learned reward function for different task settings. We also explored the effect of action spaces selection on recovering expert rewards. Benefiting from the feedback control law, we argue that using gain as action can improve the reward transfer performance.

Experiments are conducted to evaluate our approach. In the simulation, the gain-based AIRL successfully imitates the expert demonstrations during the training and has better transfer learning results than all the baselines. For the real robot experiment, although the sim-to-real gap exists, our approach successfully recovers the expert trajectory in the training and outperforms BC in a different initial position.

% While achieving an improved performance than BC, the adversarial learning framework our approach employed can be difficult to train. 
\RV{Although our approach achieves an improved generalization results by recovering the reward function of expert variable impedance skills. There still are some limitations. First, we employ a simplified impedance control law as in the previous references \cite{buchli2011learning,martin2019variable}, which doesn't measure the external force and results in a coupled Cartesian space dynamics. In the future work, we would like to decouple the dynamics by adding the external force into the impedance control law. Moreover, in our approach, we assume the waypoints are fixed and given in the task. In the future, we plan to include waypoints of the expert trajectory as one of the policy outputs and image of the workspace can be utilized as an input. In this way, our method can be extended to handle tasks with time-varying goal points.
}
% In our approach, we assume the waypoints are fixed and given in the task. For future work, we plan to include waypoints of the expert trajectory as one of the policy outputs. In this way, our method can be extended to handle tasks with time-varying goal points, and more complex trajectories can be generated.
% For the simulated Peg-in-Hole task, both the learned gain policy and reoptimized policy from learned cost function achieved $100\%$ success rate in the learning and testing scenario, while the force policy and cost function failed in the testing scenario. In the simulated Cup-On-Plate task 

\addtolength{\textheight}{-12cm}   % This command serves to balance the column lengths
                                  % on the last page of the document manually. It shortens
                                  % the textheight of the last page by a suitable amount.
                                  % This command does not take effect until the next page
                                  % so it should come on the page before the last. Make
                                  % sure that you do not shorten the textheight too much.

%%%%%%%%%%%%%%%%%%%%%%%%%%%%%%%%%%%%%%%%%%%%%%%%%%%%%%%%%%%%%%%%%%%%%%%%%%%%%%%%

%%%%%%%%%%%%%%%%%%%%%%%%%%%%%%%%%%%%%%%%%%%%%%%%%%%%%%%%%%%%%%%%%%%%%%%%%%%%%%%%

%%%%%%%%%%%%%%%%%%%%%%%%%%%%%%%%%%%%%%%%%%%%%%%%%%%%%%%%%%%%%%%%%%%%%%%%%%%%%%%%
%\section*{APPENDIX}

% \section*{ACKNOWLEDGMENT}

%%%%%%%%%%%%%%%%%%%%%%%%%%%%%%%%%%%%%%%%%%%%%%%%%%%%%%%%%%%%%%%%%%%%%%%%%%%%%%%%

\bibliographystyle{IEEEtran}
% \balance
\bibliography{IEEEabrv1}

% \begin{thebibliography}{99}
% % \bibitem{c1} Muratore, Luca, Arturo Laurenzi, and Nikos G. Tsagarakis. "A Self-Modulated Impedance Multimodal Interaction Framework for Human-Robot Collaboration." 2019 International Conference on Robotics and Automation (ICRA). IEEE, 2019.

% % \bibitem{ppo} Schulman, John, et al. "Proximal policy optimization algorithms." arXiv preprint arXiv:1707.06347 (2017).

% \end{thebibliography}
\end{document}